# Improved Stability of Whole Brain Surface Parcellation with Multi-Atlas Segmentation

Yuankai Huo*[a], Shunxing Bao[b], Prasanna Parvathaneni[a], Bennett A. Landman[a,b,c]

[a] Electrical Engineering, Vanderbilt University, Nashville, TN, USA 37235
[b] Computer Science, Vanderbilt University, Nashville, TN, USA 37235
[c] Biomedical Engineering, Vanderbilt University, Nashville, TN, USA 37235

## ABSTRACT

Whole brain segmentation and cortical surface parcellation are essential in understanding the brain's anatomical-functional relationships. Multi-atlas segmentation has been regarded as one of the leading segmentation methods for the whole brain segmentation. In our recent work, the multi-atlas technique has been adapted to surface reconstruction using a method called Multi-atlas CRUISE (MaCRUISE). The MaCRUISE method not only performed consistent volume-surface analyses but also showed advantages on robustness compared with the FreeSurfer method. However, a detailed surface parcellation was not provided by MaCRUISE, which hindered the region of interest (ROI) based analyses on surfaces. Herein, the MaCRUISE surface parcellation (MaCRUISEsp) method is proposed to perform the surface parcellation upon the inner, central and outer surfaces that are reconstructed from MaCRUISE. MaCRUISEsp parcellates inner, central and outer surfaces with 98 cortical labels respectively using a volume segmentation based surface parcellation (VSBSP), following a topological correction step. To validate the performance of MaCRUISEsp, 21 scan-rescan magnetic resonance imaging (MRI) T1 volume pairs from the Kirby21 dataset were used to perform a reproducibility analyses. MaCRUISEsp achieved 0.948 on median Dice Similarity Coefficient (DSC) for central surfaces. Meanwhile, FreeSurfer achieved 0.905 DSC for inner surfaces and 0.881 DSC for outer surfaces, while the proposed method achieved 0.929 DSC for inner surfaces and 0.835 DSC for outer surfaces. Qualitatively, the results are encouraging, but are not directly comparable as the two approaches use different definitions of cortical labels.

## 1. INTRODUCTION

Mapping the anatomical and functional relationships in the human brain is essential for image-based brain mapping. Detailed and consistent whole brain volume segmentation and surface parcellation provide the tools to establish such relationship by classifying the brain tissue and cortex into different functional regions. Many previous efforts have been proposed to perform the whole brain segmentation or surface parcellation; however, only few works provided consistent whole brain segmentation and surface parcellation [1-3]. FreeSurfer has been widely accepted as the *de facto* standard for consistent whole brain segmentation and surface parcellation using "surface-to-volume" strategy [2, 4, 5]. Recently, another "volume-to-surface" approach called multi-atlas cortical reconstruction using implicit surface evolution (MaCRUISE) was proposed to establish the consistent and robust whole brain segmentation and showed its advantages in certain aspects [6, 7]. MaCRUISE combined the multi-atlas segmentation (MAS) [8] with the Cortical Reconstruction using Implicit Surface Evolution (CRUISE) surface reconstruction [1] to achieve the consistent volume segmentation and cortical surfaces. Although it performed detailed volume segmentation (with 132 labels) and reconstructed consistent cortical surfaces, the MaCRUISE approach did not provide the cortical surface parcellatio. To understand the human anatomical and functional relationships, more regional features from cortical surfaces (e.g., area, thickness, curvature) are appealing to quantify brain anatomy for population analyses [9-11]. This work is motivated by the previous learning based surface parcellation methods [12-15].

Herein, we extend the MaCRUISE method to MaCRUISE surface parcellation (MaCRUISEsp) by developing the volume segmentation based surface parcellation (VSBSP) and topological correction functionalities (Figure 1). MaCRUISEsp has following advantages: (1) The parcellated central surface (located inside the gray matter) was provided along with the traditional inner surface (white matter surface) and outer surface (pial surface). The parcellated surfaces have been used in a recent gray matter based DTI mapping method [16]. (2) 98 cortical labels were provided by MaCRUISEsp for inner, outer and central surfaces respectively. To validate the method, 42 T1-weighted (T1w) MR volumes (21 scan-rescan longitudinal pairs from Kirby21 dataset [17]) were used. The proposed method achieved 0.94 on median Dice similarity coefficient (DSC) for central surface parcellation and superior performance on inner surface parcellation compared with FreeSurfer.

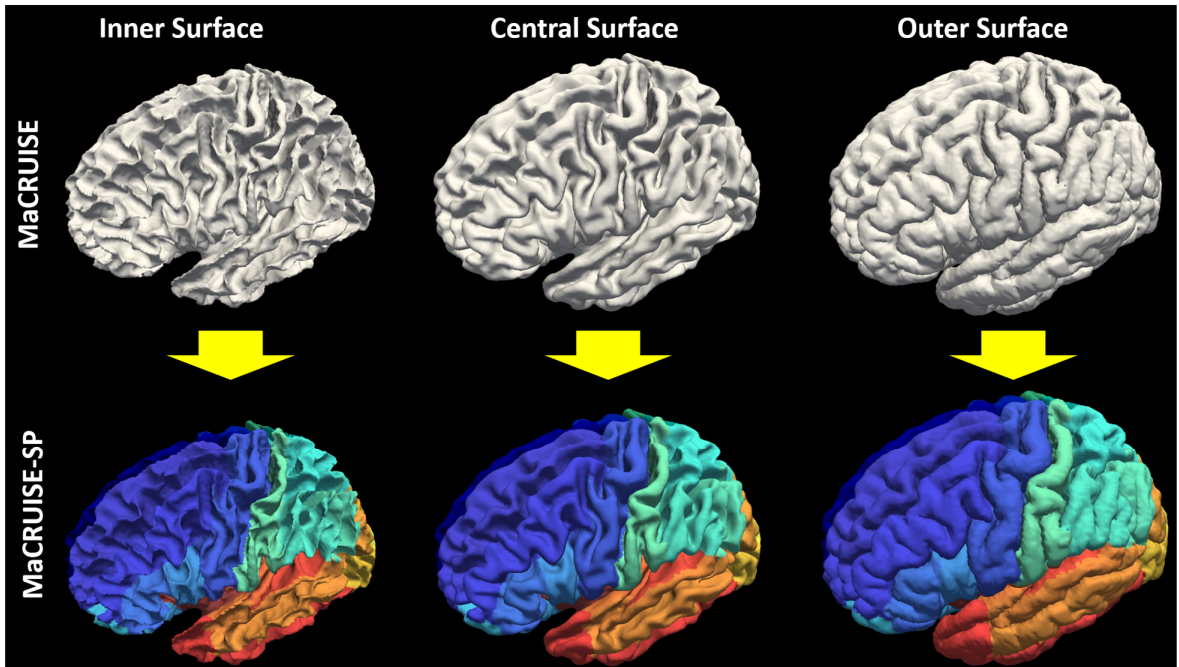

Figure 1. The motivation of MaCRUISEsp was to provide quantitative surface labels for MaCRUISE surfaces.

## 2. METHODS

**2.1 Multi-atlas Segmentation based Surface Reconstruction**

The input image of the entire processing pipeline was a single T1w brain magnetic resonance image (MRI). First, non-local spatial STAPLE (NLSS) multi-atlas segmentation framework was used to achieve whole brain segmentation[8]. Then, the MaCRUISE approach was deployed on the target image to obtain consistent whole brain segmentation and cortical surface reconstructions [6, 7]. From MaCRUISE, the inner, central, and outer surfaces were reconstructed, which were spatial consistent with volumetric segmentation (Figure 2).

**2.2 Volume Segmentation Based Surface Parcellation**

The central surface was parcellated from the whole brain volumetric segmentation. Briefly, we propagate volume labels to the central surface using the nearest label projection. For each vertex on the surface, the corresponding volumetric cortical label was assigned as the label of such vertex. This process was performed on all vertices to get the entire central surface parcellated. Since the central surface were bounded in the gray matter (GM), each vertex on the central surface were assigned a cortical label (rather than white matter or background labels). The BrainCOLOR atlas/protocol [18] was used in the proposed MaCRUISEsp framework to parcellate each surface to 98 cortical labels.

**2.3 Topological Correction**

In the BrainCOLOR protocol, each label represented a brain region with one connected component (OCC). However, after propagating the volumetric labels to surfaces, the OCC was not always ensured due to the topological mismatch. Therefore, the topological correction (TC) step was introduced to ensure each surface label to be an OCC. First, we detect the number of components of each label using "trimesh2" software (http://gfx.cs.princeton.edu/proj/trimesh2/). Then, all components on the surfaces (except the largest one) were marked as "need to fix". After repeating the previous steps for all labels, we marked all non OCC vertices as "need to fix" and fixed all of them using an iterative nearest neighbor filling strategy described in [19]. In each iteration, the remaining "needs to fix" vertices were filled by the most commonly occurring surface labels around their neighbor as the following equation:

$$\hat{L}_i = \underset{n}{\mathrm{argmax}} \sum_{k \in \gamma(i)} (L_k == n), \qquad n \in [1,2,3 \dots, N] \tag{1}$$

where $k$ indicates the indices of the labeled voxels (with the label $L_k$) around the unlabeled voxel $i$. The $n$ represents all possible $N$ cortical labels. After the topological correction, the central surface was corrected to OCC for each label.

**2.4 Surface Label Propagation**

After previous steps, the central surface was parcellated and corrected. Then, the inner and outer surfaces were parcellated by propagating the labels from the central surface. For each vertex on inner (or outer) surface, the label was propagated from another vertex on central surface, who had the smallest Euclidean distance to inner (or outer) vertex. To handle the label propagation on the back to back cortical surfaces with narrow sulcus, central vertices outside (inside) the normal plane of the vertices on the inner (outer) surfaces were considered in the distance calculation. Particularly, the nearest searching was restricted by the normal half of the plane that perpendicular to the normal direction.

## 3. EXPERIMENTS

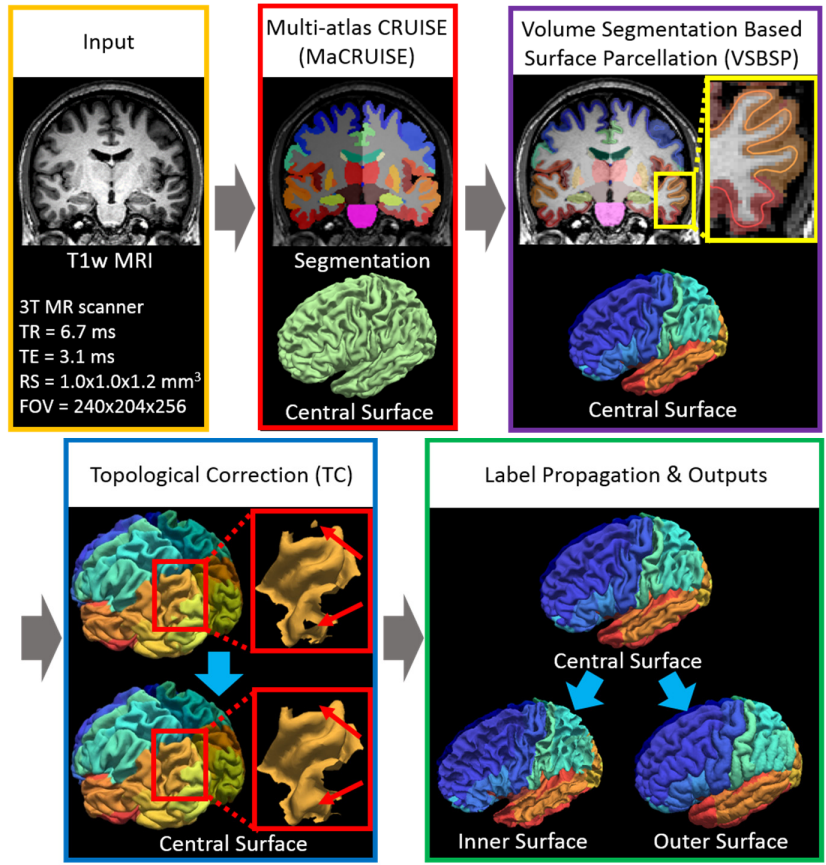

Figure 2. Work flow of MaCRUISEsp. (1) MaCRUISE was deployed on a single T1w MRI volume to achieve consistent whole brain segmentations and cortical surfaces (inner, central and outer). (2) Surface parcellation was performed on central surface using volume segmentation based surface parcellation (VSBSP). (3) The topological correction is conducted to ensure the one connected component (OCC) for each surface region. (4) The inner and outer surfaces were parcellated on by propagating the labels from central surfaces. Finally, 98 cortical labels were assigned for each surface.

## 3.1 Data

42 T1w MPRAGE MRI volumes (21 scan-rescan patients) from Kirby21 dataset [17] were used in the empirical validation to evaluate the reproducibility of the proposed MaCRUISEsp framework. The cohort consists of 11 male and 10 female patients, were collected from 3T Philips Achieva scanner with parameters: TR = 6.7 ms, TE=3.1 ms, resolution (RS) = $1.0 \times 1.0 \times 1.2mm^3$ and the field of view (FOV) = $240 \times 204 \times 256$mm.

## 3.2 Experiments

The MaCRUISEsp pipeline (Figure 2) was deployed on the dataset. Then the Dice similarity coefficient (DSC) was calculated on the parcellated scan-rescan whole brain surfaces. Briefly, each rescan surface was registered to the scan surface using rigid registration. Then the correspondence of vertices on the paired surfaces were established using the closest point matching. Finally, the DSC was derived by dividing the number of matched vertices by the average number of the vertices on the registered scan-rescan surfaces. The Wilcoxon signed rank test [20] was used for statistical analyses. All claims of statistically significance in this paper are made using the Wilcoxon signed rank test for $p < 0.01$.

## 4. RESULTS

The qualitative results (Figure 3) as well as the quantitative results (Figure 4) on the registered scan-rescan surfaces were demonstrated. In Figure 4, the reproducibility results on inner and outer surfaces using FreeSurfer Destrieux 2009 atlas were employed as the baseline performance. Note that in FreeSurfer, the Destrieux atlas has fewer labels (75 labels) on surfaces compared with the BrainCOLOR atlas (98 labels) in MaCRUISEsp framework, which would bias FreeSurfer toward larger ROIs and higher DSC. The Pearson correlation results (surface area and cortical thickness) across 21 scan-rescan pairs for all cortical labels were provided in the Figure 5.

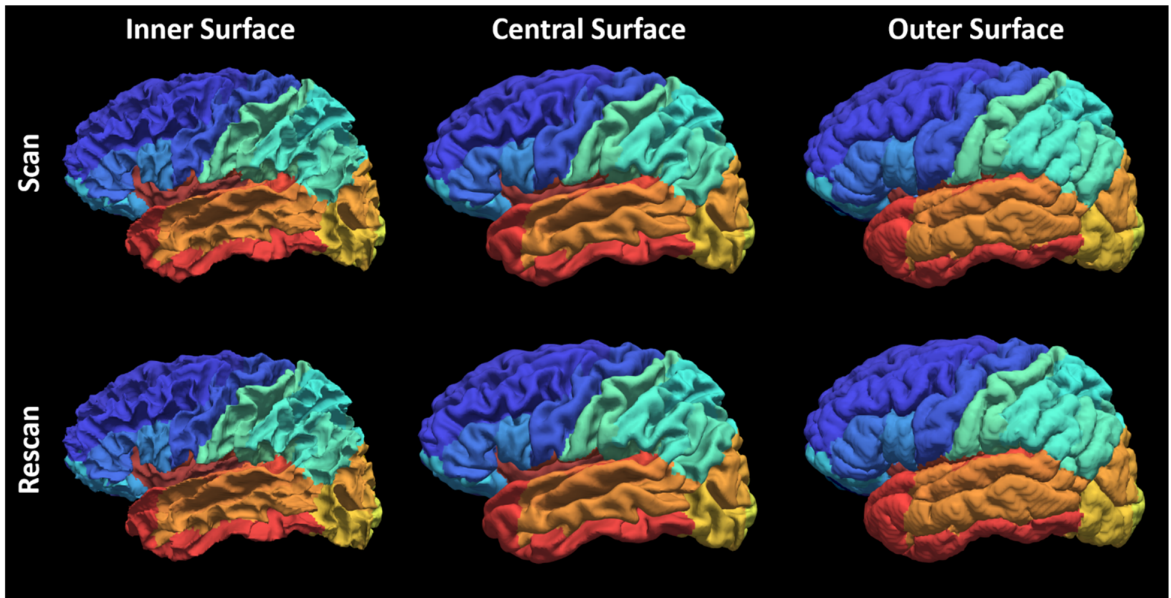

Figure 3. Qualitative reproducibility results on the surface parcellation between a randomly selected scan-rescan patient using MaCRUISEsp.

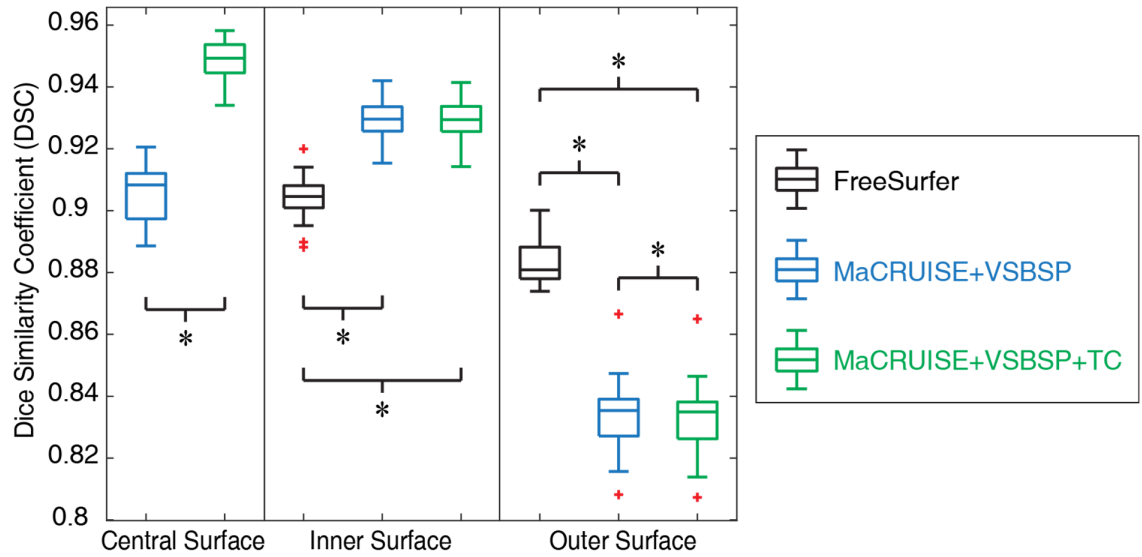

Figure 4. Quantitative segmentations results on the surface parcellation for the entire Kirby21 cohort. The reproducibility on inner and outer surfaces using FreeSurfer's Destrieux 2009 atlas (75 labels) were employed as the baseline. The MaCRUISE+VSBSP method as well as the MaCRUISEsp (MaCRUISE+VSBSP+TC) method using BrainCOLOR atlas (98 labels) were presented. The symbol "*" indicated the differences are significant for the Wilcoxon signed rank test for $p < 0.01$.

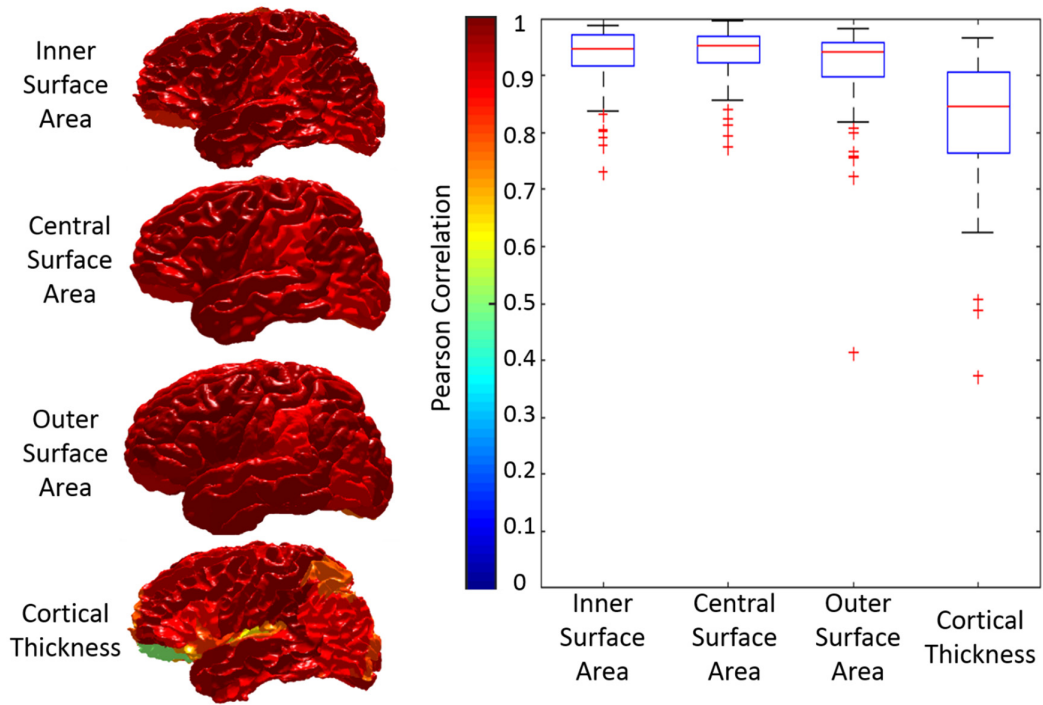

Figure 5. The reproducibility of surface metrics (surface area and cortical thickness) were shown. The Pearson correlation values for four metrics on each label were shown in the left panel. The color of each label corresponds to the Pearson correlation value showed in the color bar. Then, the qualitative results of all labels were shown as the boxplot in the right panel.

## 5. CONCLUSION AND DISCUSSION

We present the MaCRUISEsp method for the whole brain surface parcellation. From the experimental results, the "volume-to-surface" strategy with topological correction provided us 0.95, 0.92 and 0.85 median DSC for central surface, inner surface and outer surface respectively (Figure 4). The results showed that the MaCRUISEsp provided the central surface parcellation, which not was typically provided by FreeSurfer. With topological correction, the MaCRUISEsp obtained the generally better reproducibility than without using topological correction. The proposed methods achieved significantly higher reproducibility than FreeSurfer on inner surface parcellation while the FreeSurfer achieved significantly higher reproducibility than the proposed methods on the outer surface parcellation. Note that the comparison was made in the situation that more labels were provided by MaCRUISEsp (98 labels) compared with FreeSurfer (75 labels). For a more thoughtful analysis, the reproducibility on the surface metrics were provided in Figure 5. Qualitatively, the results from proposed methods were encouraging, but are not directly comparable to FreeSurfer as the two approaches use different definitions of cortical labels.

## 6. ACKNOWLEGEMENTS

This research was supported by NSF CAREER 1452485, NIH grants 5R21EY024036, 1R21NS064534, 2R01EB006136 (Dawant), 1R03EB012461 (Landman) and R01NS095291 (Dawant). InCyte Corporation (Abramson/Landman). This research was conducted with the support from Intramural Research Program, National Institute on Aging, NIH. This study was in part using the resources of the Advanced Computing Center for Research and Education (ACCRE) at Vanderbilt University, Nashville, TN. This project was supported in part by ViSE/VICTR VR3029 and the National Center for Research Resources, Grant UL1 RR024975-01, and is now at the National Center for Advancing Translational Sciences, Grant 2 UL1 TR000445-06. We are grateful for the assistance of Kunal Nabar in helping to prepare this manuscript. We gratefully acknowledge the support of NVIDIA Corporation with the donation of the Titan X Pascal GPU used for this research.